\documentclass[letterpaper, 10 pt, conference]{ieeeconf}  

\IEEEoverridecommandlockouts                              

\usepackage{graphics} 
\usepackage{epsfig} 
\usepackage{mathptmx} 
\usepackage{times} 
\usepackage{amsmath} 
\usepackage{amssymb}  
\usepackage{cite}

\usepackage{bookmark}

\usepackage{amsfonts}
\usepackage{algorithmic}
\usepackage{graphicx}
\usepackage{subcaption}
\usepackage{textcomp}
\usepackage{array}
\usepackage{xcolor}
\usepackage{layouts}
\usepackage{siunitx}

\title{\LARGE \bf
DeepAerialMapper: Deep Learning-based Semi-automatic HD Map Creation for Highly Automated Vehicles
}

\usepackage{xprintlen}

\author{Robert Krajewski$^{1}$ and Huijo Kim$^{2}$
\thanks{$^{1}$Robert Krajewski is with RWTH Aachen University,
        {\tt\small robert.krajewski@rwth-aachen.de}}%
\thanks{$^{2}$Huijo Kim is with hexafarms,
        {\tt\small ccomkhj@gmail.com}}%
}

\begin{document}
\let\oldtwocolumn\twocolumn
\renewcommand\twocolumn[1][]{%
\oldtwocolumn[{#1}{
\begin{center}
\centerline{\includegraphics[width=\textwidth]{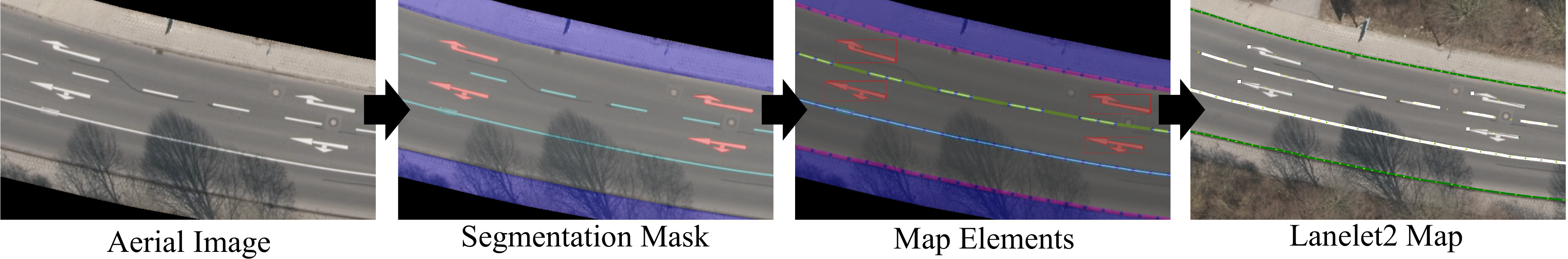}}
\captionof{figure}{Overview of the individual automatic processing steps from aerial image to HD map. Using a neural network, the image is segmented semantically into the relevant classes (see Table~\ref{table:class_distribution}). The individual elements, including lane markings, road borders and road arrows, are then detected in the semantic mask, classified and grouped. In the last step, the results are exported in Lanelet2 format. Subsequently, the map can be manually enhanced, e.g., with non-visible complementary information such as speed limits or optional elements like centre lines.}
\label{fig:overview}
\end{center}
}]
}

\maketitle
\thispagestyle{empty}
\pagestyle{empty}

\begin{abstract}

High-definition maps (HD maps) play a crucial role in the development, safety validation, and operation of highly automated vehicles. Efficiently collecting up-to-date sensor data from road segments and obtaining accurate maps from these are key challenges in HD map creation. Commonly used methods, such as dedicated measurement vehicles and crowd-sourced data from series vehicles, often face limitations in commercial viability. Although high-resolution aerial imagery offers a cost-effective or even free alternative, it requires significant manual effort and time to transform it into maps. In this paper, we introduce a semi-automatic method for creating HD maps from high-resolution aerial imagery. Our method involves training neural networks to semantically segment aerial images into classes relevant to HD maps. The resulting segmentation is then hierarchically post-processed to generate a prototypical HD map of visible road elements. Exporting the map to the Lanelet2 format allows easy extension for different use cases using standard tools. To train and evaluate our method, we created a dataset using public aerial imagery of urban road segments in Germany. In our evaluation, we achieved an automatic mapping of lane markings and road borders with a recall and precision exceeding 96\%. The source code for our method is publicly available at https://github.com/RobertKrajewski/DeepAerialMapper.

\end{abstract}

\section{INTRODUCTION}

Current trends in electromobility and automated driving will shape the future of transportation. Although the share of electrified vehicles is already substantial, automated vehicles at higher levels of automation are still either prototypes or restricted to specific geographical areas. The development, safety validation and operation of highly automated vehicles pose a complex problem, as the human driver's task must (temporarily) be completely taken over depending on the automation level \cite{sae}. The driving task itself can be divided into different components, including perceiving surrounding traffic, interpreting the current driving scenario, and planning the vehicle's actions.

\begin{figure*}[t!]
    \includegraphics[width=\textwidth]{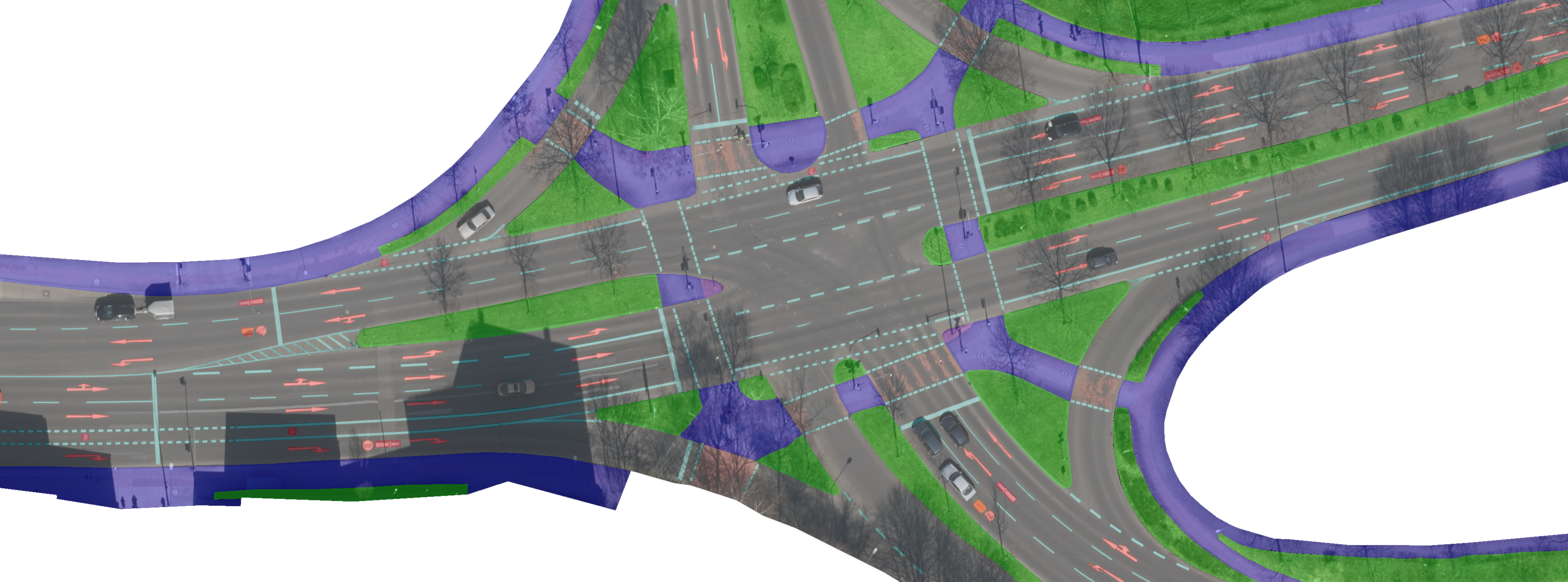}
       \captionof{figure}{Exemplary semantic segmentation of an intersection shown as overlay. The annotated classes are represented by colours (Red: symbols, cyan: lane markings, gray: road, green: vegetation, blue: walkway, purple: traffic island)}
       \label{fig:headliner}
\end{figure*}

For some of these tasks, automated vehicles benefit from supplementing their perception and positioning systems by map information~\cite{liu_wang_zhang_2020}. While standard-definition (SD) maps describing the coarse road topology are sufficient for, e.g., route planning, higher levels of automation require high-definition maps. Detailed lane-level information, including lane markings and symbols, can be used by various subsystems of automated vehicles: One example is localisation, which can better estimate the vehicle's position based on known prominent points in the immediate environment \cite{s18103270, poggenhans2018precise}. Another example is behaviour planning, as vehicle-mounted sensors perceive on limited ranges and suffer from vehicle-to-vehicle occlusions. HD maps extend the visual horizon beyond the range of installed sensors, enabling a more predictive driving style, e.g., getting into the correct lane at an early opportunity \cite{yoon2021high}. Safety is also enhanced, for example, by facilitating the perception to more easily differentiate between traffic signals and a car's rear lights at intersections~\cite{yoneda2020robust}.

When creating and using HD maps, it is essential to ensure that the resulting maps are complete, correct, and up-to-date.
The methods used to collect the raw data and process it into an HD map determine how this can be achieved and at what cost.
There are two fundamental methods for creating data for HD maps. The typical method is recording road segments from a vehicle's perspective. Dedicated measurement vehicles with extensive sensor setups are used to perform systematic measurement campaigns~\cite{liu_wang_zhang_2020}. In addition, series vehicles with cheaper sensor setups and data-upload capabilities are used to continuously crowd-source data \cite{crowd}. However, as access to dedicated measurement vehicles or a fleet of series vehicles is required, these methods are mainly used by car manufacturers or companies specialising in mapping. 
The alternative to the vehicle perspective is the aerial perspective (e.g., aircraft or satellites), which has received less attention to date. 
Due to the typically low resolution in terms of Ground Sampling Distance (GSD), most research has focused on producing SD maps that only cover road topology \cite{xu2021topo}. However, resolution is steadily increasing, and data is becoming more affordable and available, as, for example, local authorities provide digital orthophotos free of charge. 
Although the orthophotos have already been pre-processed and geo-referenced, one issue is that the manual creation of the HD map is labour-intensive. For instance, the boundaries of each lane must be manually annotated with polylines and attribute tags in mapping software such as JOSM \cite{josm}.

Inspired by existing systems for the automated creation of prototypes for significantly simpler SD maps \cite{vargas2020openstreetmap}, we have developed a method to support the creation of HD maps.
Although methods for semantic segmentation of aerial imagery already exist, they lack the necessary complex post-processing and export to an HD map.
As a foundation for our method, we create a dataset of freely available, high-resolution, geo-referenced aerial images of urban road segments in Aachen, Germany.
We manually semantically annotate these using eight categories in order to train a convolutional neural network for semantic segmentation.
To subsequently create an HD map from a semantic segmentation, we present a method that hierarchically detects, groups, classifies, and combines individual elements such as symbols and lane markings.
For this purpose, mainly classical methods for image processing are used, as well as a separate neural network for symbol classification. 
The result of these automatic processing steps for the visible road elements is a prototypical HD map, which already covers the otherwise very labour-intensive elements.
As our method is purely vision-based, we export the resulting map in the Lanelet2 format \cite{lanelet2} to allow manual completion of the map using standard tools.
This is necessary since, typically, some map elements are barely visible or occluded by a vehicle or tree. 
In addition, other elements, such as speed limits or signs, cannot be extracted from aerial imagery and therefore must be added manually from complementary sources.

\section{RELATED WORK}

\subsection{Data Acquisition}

Both the vehicle and aerial perspectives are viable options for collecting imagery for HD maps. Depending on the method and perspective chosen, there are advantages and disadvantages in terms of costs, accuracy, processing effort, and result verification.

Dedicated measurement vehicles are used to capture road data from a vehicle's perspective. These vehicles are typically equipped with a high-precision inertial navigation system for localisation, along with multiple camera and lidar sensors to record their surroundings. This setup allows for high-quality data capture but it necessitates a complex pre-processing pipeline to fuse the data from individual sensors and temporally aggregate the perceived information. This is crucial for handling temporary occlusions caused by other objects or vehicles that block the line of sight. However, expensive sensor setups impose limitations on the scale of the vehicle fleet used for data recording in measurement campaigns~\cite{racq2020}.

An alternative method involves crowd-sourcing data by employing a fleet of series vehicles that upload data during regular operations~\cite{liu2020creating}. With a large fleet, it becomes easier to collect substantial amounts of data and keep maps of frequently used road segments up-to-date. Consequently, no dedicated measurement campaigns are required to cover changes in symbols or road construction. However, the deployment of large fleets of vehicles is typically feasible only for car manufacturers, and series vehicles often utilise simpler GPS modules and less sophisticated sensor setups compared to measurement vehicles. Having to partially process the collected data online within the vehicle, primarily for data privacy and to manage the uploaded data volume, further compromises its quality. Hence, only processed and anonymised data, such as detected sign positions or lane markings, are typically uploaded~\cite{crowds}. Since raw sensor data is not available for map creation, utilising more sophisticated object detection methods or verifying the sent data becomes impractical. Consequently, the uploaded data must undergo extensive post-processing to extract an HD map from the noisy data collected by individual vehicles.

An alternative to vehicle-based data collection is to acquire data from an aerial perspective. Aerial images can be efficiently captured using aircraft, satellites, or drones. The resulting resolution, known as the ground sampling distance, depends on factors such as flight altitude and camera optics. Currently, aircraft and drones can achieve resolutions of around \SI{5}{\centi\metre\per{px}} or higher, while civil satellites are limited to approximately \SI{25}{\centi\metre\per{px}}~\cite{8768011}. While aerial perspective enables coverage of large areas, certain objects, such as signs, may not be visible or can be occluded by large trees. One significant advantage of aerial imagery is its accessibility, as it is often created by local governments for various purposes like surveying, preprocessed, and occasionally made freely available as digital orthophotos on a regular, annual basis.

\subsection{Processing Aerial Imagery}

The processing of static aerial imagery has numerous potential applications across various domains, including agriculture and municipal use. In this paper, we focus specifically on methods for generating road maps. These methods can be categorised based on whether they create Standard Definition (SD) or High Definition (HD) maps. For road-level maps (SD), aerial imagery with lower resolution is generally sufficient. For instance, in \cite{c3} and \cite{deeproadmapper}, data-driven methods have been proposed for detecting roads and determining their connectivity.

Other methods utilise higher-resolution imagery to extract the elements required for the creation of HD maps, with a particular emphasis on lane markings.

Kim et al. \cite{kim2006efficient} propose an efficient method for detecting lane markings and symbols in complex urban regions using a set of simple image processing algorithms, including template matching. Azimi et al. \cite{azimi2018aerial} propose a method based on deep learning, where the typical encoder-decoder architecture is enhanced by wavelet transforms to achieve accurate semantic segmentation of lane markings. In \cite{markscaps}, a network architecture dedicated to semantic segmentation of lane markings is proposed, utilising capsule networks and self-attention.

Although existing methods have focused on accurately extracting high-level features, such as road connectivity, or low-level features, such as lane markings, from aerial images, none of them provide a complete semantic segmentation of a road segment in an aerial image to subsequently create a detailed HD map, including lane markings and symbols, in a typical format.

In summary, the existing research has primarily concentrated on extracting specific road features, such as road connectivity or lane markings, from aerial images. However, there is a lack of methods that generate a comprehensive semantic segmentation of an aerial image, encompassing all relevant road features, including symbols, lane markings, roads, vegetation, walkways, and traffic islands, to create an HD map.

\subsection{Lanelet2 Format}

There are several formats available for HD maps, each offering different features and tools. One prominent format is Lanelet2~\cite{lanelet2}, an open-source format introduced in 2018 that is primarily used in research. Lanelet2 uses a hierarchical structure to define the various components of an HD map. At the lowest level, the map consists of points (nodes) that represent positions in global coordinates. These points are organised into lines (linestrings), which are used to describe lane markings and other similar features. These lines can then be combined to create lanelets, which represent short segments of a lane. Lanelets can also be annotated with symbols and regulatory elements, such as speed limits. The storage format for Lanelet2 is an extension of the XML-based OSM file format. Given its intuitive nature, popularity within the research community and compatibility with free editing tools, we have selected the Lanelet2 format as the basis for our method. As the Lanelet2 format encompasses a vast range of elements, this paper focuses specifically on the essential elements that can be extracted from an aerial perspective.

\section{SEMANTIC SEGMENTATION OF AERIAL IMAGERY}

High-resolution aerial imagery, captured by satellites or aircraft, is typically available as tiles of RGB images. In order to analyse the composition of the visible road segment at each pixel position, semantic segmentation is required. While each pixel in the RGB image represents a colour, semantic segmentation assigns a category from a predefined catalogue of semantic categories to each pixel, based on the object visible at that location. This representation is necessary for the subsequent process of creating the HD map.

Deep neural networks are the state-of-the-art method for generating semantic segmentations. These networks require training data to learn the mapping between RGB images and semantic segmentation. Specialised neural network architectures have been proposed for aerial imagery; however, since our focus is primarily on map generation, we utilise less complex architectures and simpler semantic categories, which have proven to be sufficient.

At the time of this paper's inception, there was no publicly available dataset that provided both sufficiently high resolution and all the necessary categories. Hence, we created our own dataset. We used aerial images published by the Aachen city administration, which are freely available for research purposes. We selected a total of 63 digital orthophotos taken in and around Aachen, Germany. These images depict typical urban elements such as straight roads, intersections, and roundabouts. All images have been rectified and geo-referenced, enabling the determination of each pixel's position in UTM coordinates. Additionally, the images have a ground-sampling distance of 5 centimetres per pixel, providing a sufficiently high resolution to clearly discern individual elements such as lane markings. Typically, the images cover an area of approximately \qtyproduct{180x70}{\metre}.

In order to reduce the manual annotation effort and consider the subsequent steps in the process, we removed all areas irrelevant to an HD map from the images in the initial annotation step by painting them over in black. This includes areas that are not in close proximity to a road or walkway. We used eight additional classes, divided into three groups, to annotate the remaining image areas. The first group describes the ground surface and includes five classes: road, walkway, parking space, vegetation, and traffic island. The second group comprises ground markings, such as lane markings and symbols. Symbols primarily encompass road arrows but can also include text written on the road. The last group involves annotating vehicles to detect occlusions of symbols, although this group is not used for semantic segmentation. Annotation was performed group by group, starting with the first group and not considering objects from subsequent groups. For example, a street's annotation disregards occlusions caused by vehicles.

Limiting the annotations to lane markings and symbols, as proposed in \cite{azimi2018aerial}, would have considerably reduced the annotation effort. However, this approach would have been impractical due to the absence of crucial information necessary for creating an HD map. For instance, the absence of lane markings on the outer edge of some roads implies that the transition from the road to vegetation only implicitly describes the end of a lane.

Figure~\ref{fig:headliner} shows an annotation overlay on the RGB image. The class distribution (see Table~\ref{table:class_distribution}) reveals a notable imbalance among the classes. Particularly relevant classes, such as lane markings and symbols, are noticeably underrepresented. Although more than 50\% of the total annotation points are used for annotating these polygons, they only cover less than 1.5\% of the total area. Consequently, the weight of these classes was significantly increased during training. Another challenge arises from the fact that objects like lane markings or symbols are very small despite the high ground-sampling distance, yet they need to be accurately segmented, especially in the case of road arrows. Therefore, we chose network architectures capable of providing non-upscaled semantic segmentation at full or half input resolution.

\begin{table}[htbp]
\caption{Dataset Class Distribution}
\label{table:class_distribution}
\begin{center}
\begin{tabular}{cccc}
\hline
\textbf{Irrelevant} & \textbf{Road} & \textbf{Walkway} & \textbf{Vegetation}    \\
\hline
58.7\% & 14.8\% & 5.22\% & 19.9\% \\
\hline
\hline
\textbf{Parking} & \textbf{Traffic Island} & \textbf{Symbol} & \textbf{Lane Marking} \\
\hline
0.210\% & 0.214\% & 0.109\% & 0.826\% \\
\hline
\end{tabular}
\label{tab1}
\end{center}
\end{table}

\section{SYMBOL CLASSIFICATION}

In addition to lane markings, symbols play a crucial role in HD maps for vehicle trajectory planning. Symbols provide directional information for vehicles.
Given the presence of various symbol categories and their relatively small size in aerial images, it is beneficial to separate their classification from semantic segmentation by employing a dedicated network.
Among the symbols in our dataset, road arrows are the most frequently observed. Therefore, this paper focuses on road arrows and classifies them based on their pointing directions.
As the classification relies solely on shape rather than colour or texture, we can classify the symbols using their mask in the semantic segmentation.
If less than half of a road arrow is misclassified as a lane marking, potentially due to the same colour, we automatically detect this and assign the symbol class to all connected pixels.
Following that, we extract the road arrow's mask from the resulting semantic segmentation for further classification.
Given the standardised shape of the symbols, their classification is relatively straightforward compared to the semantic segmentation of the entire image.
However, our experiments demonstrated that traditional classification methods, such as utilising HuMoments~\cite{6308139}, did not yield satisfactory results. Consequently, we developed a dedicated neural network for this purpose.

\section{CREATION OF HD MAPS}

The creation of HD maps is based on the semantic segmentation of aerial images and symbol classification. The segmentation identifies relevant parts in the image, which may be grouped and abstracted as needed to be mapped in the Lanelet2 format. Classical image processing techniques are primarily employed for this purpose. The outcome is an automatically generated prototypical HD map that encompasses visible elements and can subsequently be subjected to manual refinement using complementary information.

The process comprises a series of sequential steps. Initially, low-level elements such as contours are extracted through segmentation. After that, these elements are progressively processed and combined to derive road borders, lane markings, and lanes. The following overview provides a brief description of each step:

\begin{enumerate}
\item Segmentation: Aerial images are subjected to semantic segmentation to identify and isolate relevant parts.

\item Extraction of low-level elements: Initially, contours are extracted as low-level elements from the segmented image.

\item Processing and combination: The extracted contours are further processed and gradually combined to obtain road borders, lane markings, and eventually lanes.

\item Lanelet2 format mapping: The abstracted elements are mapped in the Lanelet2 format, facilitating their integration into the HD map.
\end{enumerate}

The above steps provide a high-level overview of the HD map creation process, with further implementation details available in the associated source code publication.

\subsection{Road Border}

The initial processing step involves extracting the drivable area and its boundary from the semantic segmentation. To achieve this, a mask is generated that includes all pixels belonging to the classes of road, lane markings, and symbols. Subsequently, the boundary of this area is determined as a set of contours. Since these contours describe the boundary with pixel-level precision and may exhibit slight noise, they are approximated as polylines through downsampling, enabling simplified further processing.

\begin{figure}%
    \centering
    \subfloat[\centering Before splitting]{{\includegraphics[width=3.8cm]{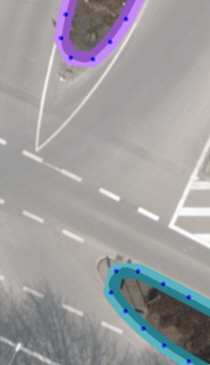} }}%
    \qquad
    \subfloat[\centering After splitting
    ]{{\includegraphics[width=3.8cm]{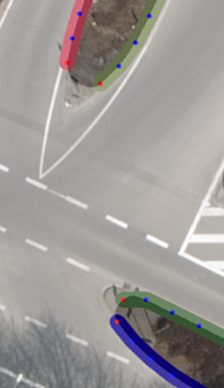} }}%
    \caption{Exemplary result of road border splitting. Different colours represent individual road borders.}%
    \label{fig:split}%
\end{figure}

To ensure compliance with the Lanelet2 format, the polylines are then split whenever consecutive segments exhibit significant differences in orientation. This allows the generation of separate contours for each driving direction, as exemplified in Fig.~\ref{fig:split}.

In the final stage, the contours are classified according to the Lanelet2 format~\cite{lanelet2} as either road borders or curbstones. This classification is based on the presence or absence of a sidewalk adjacent to the contour in the semantic segmentation.

\subsection{Lane Markings}

Once the drivable area is identified, all lane markings within it are located and classified. The process for this is similar to that for the road border, as a mask and contours are also extracted from the semantic segmentation. However, since the contours of the lane markings are not of interest, only their centre lines are extracted from the skeletonised mask as contours (see Fig.~\ref{fig:overview}). Another important characteristic of lane markings is that they can split, for example, when lanes divide, and they can also merge subsequently. To detect these scenarios, the number of neighbouring points for each point on the contour is calculated. If a point has three neighbours, the angled part is identified as a separate contour. Detecting \ang{90} angles, like those at stop lines, involves a similar process. 
The final step involves grouping the dashed lane markings. Initially, the lane markings are detected in the mask by a size filter. Subsequently, they are iteratively grouped by searching for similar lane markings in terms of length and orientation. In the end, the lane markings are classified as either \emph{dashed} or \emph{solid}.

\subsection{Lanes}

The final processing step heuristically pairs the extracted information to form lanes. Road borders and lane markings are available as classified polylines resulting from the previous steps. To perform the pairing, possible corresponding counterparts are searched for each segment of the road borders or lane markings. We accomplish this by taking into account the distances and orientations of the elements. If a sufficient number of matches are found for the segments and there is a continuous road between the elements (without any obstructions such as a traffic island), they are assigned as part of a lane.

\subsection{Exporting as Lanelet2 Map}

The final map needs to be exported in Lanelet2 format for further processing and utilisation. To accomplish this, all elements are converted from the image coordinate system to UTM coordinates. This conversion requires knowledge of the position of the upper left corner and the ground sampling distance for each image. Upon exporting, the internal map elements are mapped to the elements in Lanelet2 format, which utilises and extends the OSM format. Consequently, points are represented as OSM nodes, linestrings (polylines) are transformed into ways, and lanelets are converted into relations. Two points along the major axis and the symbol category describe detected symbols.

\begin{figure}[t!]
\centerline{\includegraphics[width=\columnwidth]{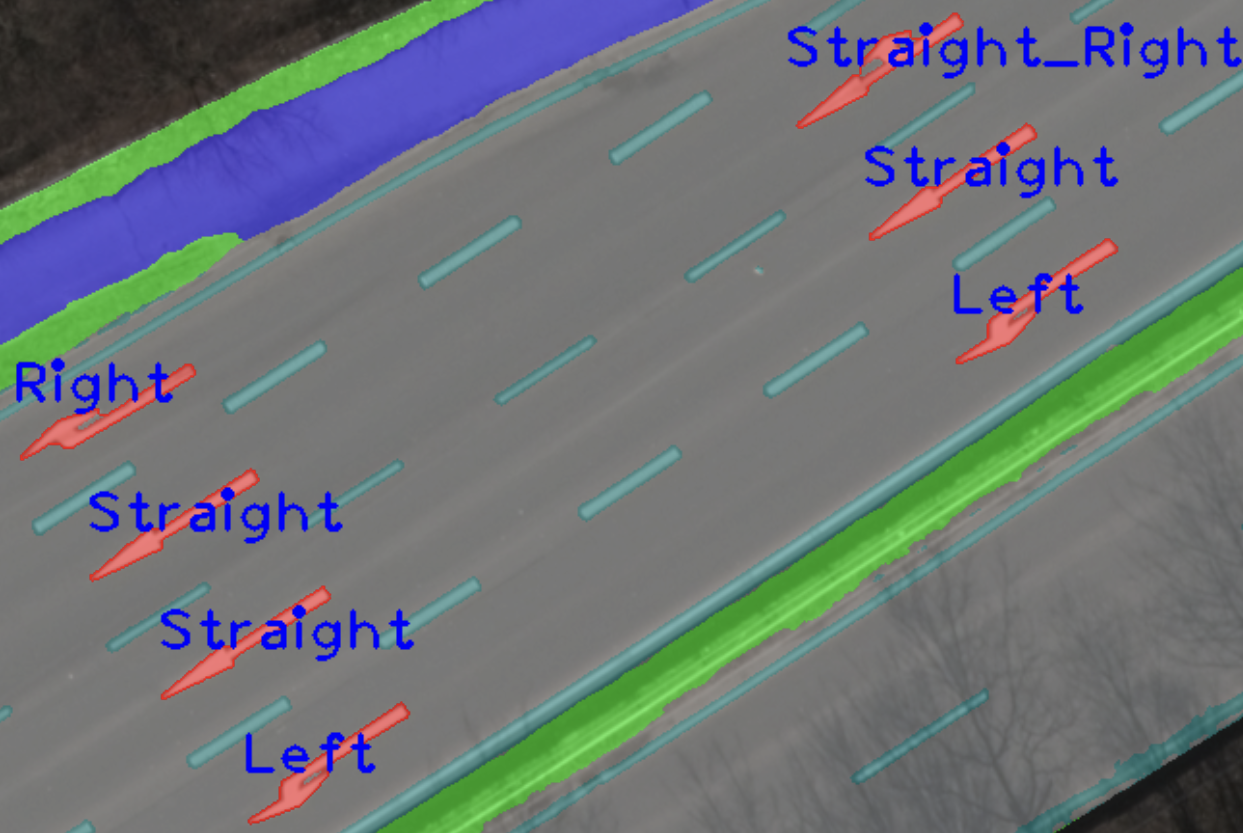}}
\caption{Exemplary result of symbol classification.}
\label{fig:symbol_classification}
\end{figure}

\section{EXPERIMENTS}

For a comprehensive evaluation of the resulting system, we evaluate semantic segmentation, symbol classification, and map creation as separate components.

\subsection{Semantic Segmentation}

As our initial architecture, we select the UNet \cite{c10} architecture, which is well-suited for segmenting small objects since it operates at full resolution. However, due to high memory requirements, we modify the architecture by removing the last upsampling and the last block in the decoder to enable prediction at half resolution. As an alternative architecture, we employ a simplified UPerNet \cite{lin2017feature} architecture, which combines FPN \cite{c11} with PPM \cite{c13} to achieve a larger receptive field. To obtain a semantic segmentation at half the original resolution, we modify the variant by eliminating the maxpooling layer at the beginning of the ResNet with a ResNet50 \cite{c14} backbone for prediction.

To train the semantic segmentation networks, we randomly divide the entire dataset into training, validation, and testing sets in a 70:15:15 ratio. We train all networks using SGD with a learning rate of 0.01. Since the aerial images are large, we train on random crops and apply additional augmentation techniques such as rotations, scaling, and mirroring. Given the under-representation of lane markings and symbols in the cross-entropy loss, we empirically set the class weights for these two classes to 40 and 20, respectively. Table~\ref{table:semantic_segmentation} presents a comparison of networks that use mIoU and IoU for the symbol class, which is crucial for the symbol classifier. The results indicate that the segmentation resolution used does not significantly affect the mIoU. However, upon qualitative examination of the segmentation masks, it is evident that only at half or full resolution is the segmentation of symbols sufficiently detailed for subsequent classification. 
A more substantial quantitative difference is observed when using different architectures, with the UPerNet architecture yielding significantly better results. Furthermore, the results exhibit less noise, likely due to the larger receptive field. One limitation of all networks is the poor segmentation of parking areas and traffic islands. However, even for human annotators, distinguishing these areas during the annotation process was challenging, and the data only sparsely represents them (see Section III). Fortunately, these classes are less critical for HD map creation than the classes used to derive contours (lane markings, symbols, and roads).

\begin{table}[htbp]
\caption{Semantic Segmentation Results}
\label{table:semantic_segmentation}
\begin{center}
\begin{tabular}{cc|cc}
\hline
\textbf{Architecture} & \textbf{Resolution} & \textbf{mIoU} & \textbf{IoU[Symbol]} \\
\hline
UNet & full & 0.64 & 0.74 \\
UNet & half & 0.65 & 0.76 \\
UPerNet & half & \textbf{0.70} & \textbf{0.80} \\
UPerNet & quarter & 0.69 & 0.76  \\
\hline
\end{tabular}
\label{tab2}
\end{center}
\end{table}

\subsection{Symbol Classification}

\begin{figure}[h!]
    \centering
    \subfloat[\centering Vehicles occluding symbols.]{{\includegraphics[width=4.8cm]{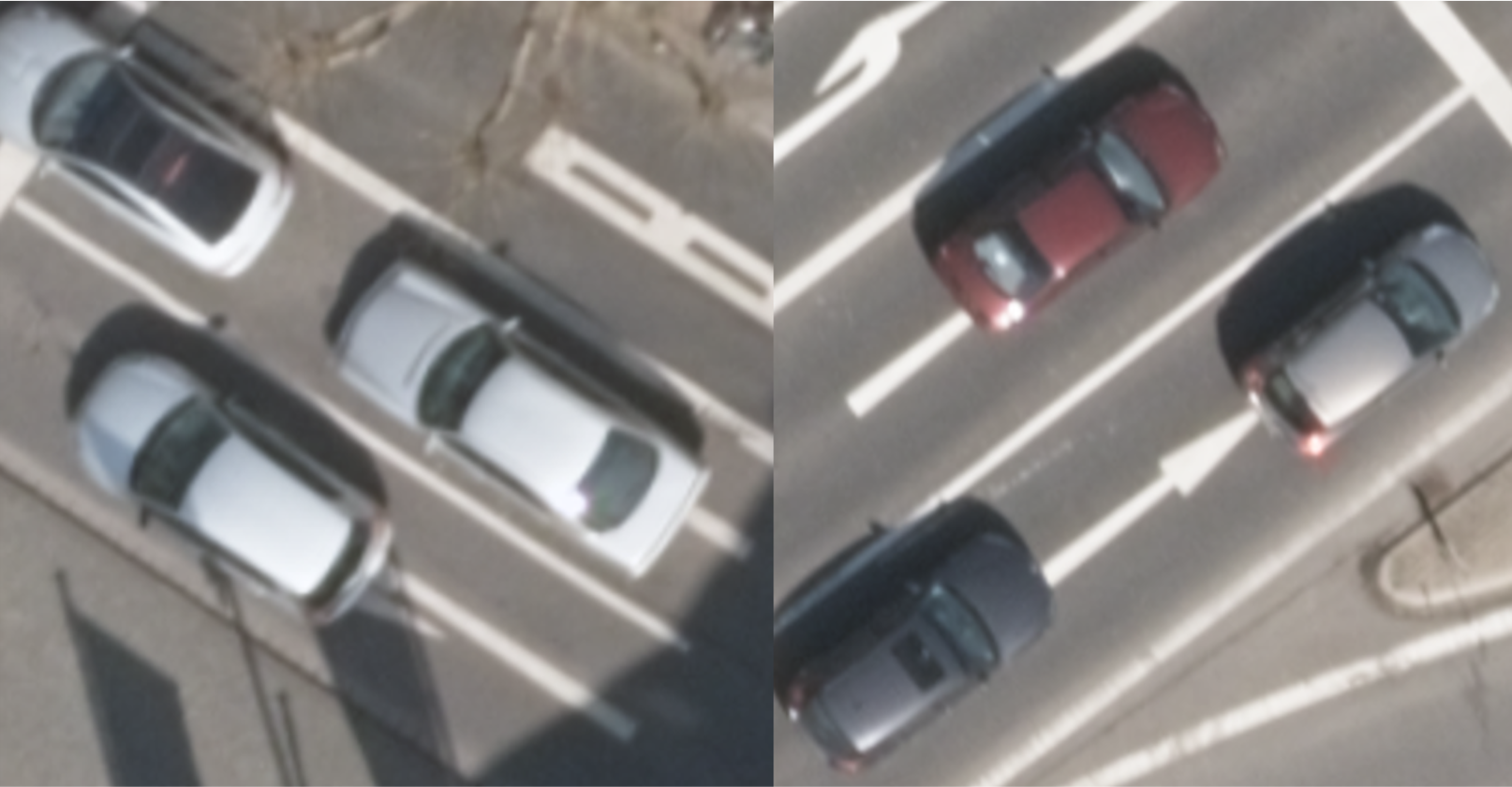} }}%
    \qquad
    \subfloat[\centering Simulated occlusions.]{{\includegraphics[width=2.6cm]{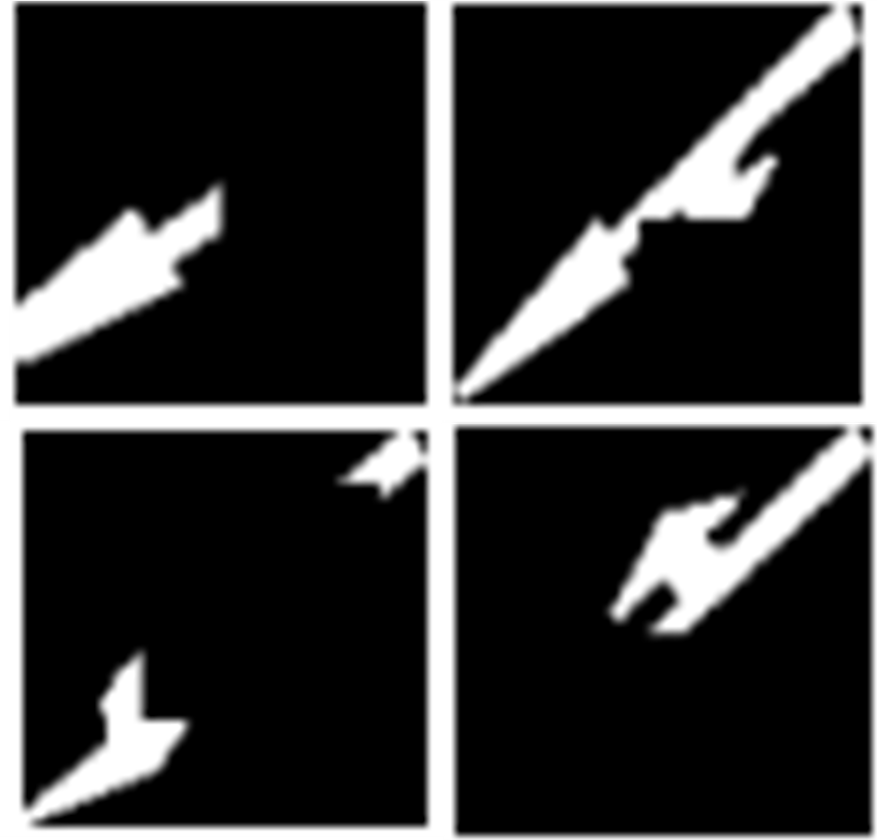} }}%
    \caption{Mimicking occlusion of symbols by cropping and cutting out random parts.}
    \label{fig:sym_augmentation}%
\end{figure}

For the purpose of training, validation, and testing, we utilised symbol masks extracted from the corresponding semantic segmentations produced by one of our trained networks. In order to expand the training and validation sets, which initially consisted of 470 symbols, we applied augmentation techniques such as slight rotation and scaling to the images. Since road arrows in aerial imagery are frequently subject to occlusion by vehicles and other objects, or may be cropped at the image boundaries, we also introduced random cropping and mask deletion to simulate these scenarios \cite{devries2017improved} (refer to Fig.~\ref{fig:sym_augmentation}). Our symbol classification involves seven defined classes, namely left, right, straight, straight-or-left, straight-or-right, left-or-right, and other. The "other" class encompasses symbols such as symbols at bus stops. To evaluate the robustness of our trained network against rotations, we augmented the test set with randomly rotated instances, ranging from 48 to 192 symbols.

For our network architecture, we adopted a simple sequential model consisting of five convolutional layers with ReLU activations. The network was trained using Stochastic Gradient Descent (SGD) with a learning rate of 0.01 (gamma=0.92) and a batch size of 64, until convergence was achieved on the validation dataset. When evaluating the trained network on the test set, we observed a high accuracy of 94\%. While the segmentation network failed to accurately segment two symbols, partial occlusions were responsible for the remaining misclassifications.

\subsection{Map Creation}

For the evaluation of the automatically created maps, manually created maps are necessary as a reference. Since the manual annotation of road borders and lane markings by polylines typically takes a large part of the time, we focused our evaluation on this. Accordingly, we manually annotated all lane markings and road borders by polylines for all images in the test dataset.

In order to evaluate the automatically created polylines, their completeness and their local precision are to be analysed based on their control points. Since the control points of the automatically and manually created polylines are inhomogeneously distributed, they are resampled uniformly with a distance of \SI{10}{\centi\metre}. Subsequently, individual points can be optimally assigned in a 1:1 manner, regardless of class, using the Hungarian algorithm with a maximum assignment distance of \SI{20}{\centi\metre}. We calculate the precision and recall for the test dataset based on the resulting assignments.

To determine the impact of semantic segmentation errors on the generated map, we created maps for both the manually annotated semantic segmentation and the semantic segmentation created by the trained network. The results for the test set (see Table~\ref{table:map_error}) show that the polylines are positioned with a high recall of 97\% on the ground truth data. A high recall is essential, as it largely eliminates the need for time-consuming manual annotation of the contours. At the same time, there are only a few false positives, resulting in a precision of 97\%. Visually analyzing the results reveals that errors typically occur near the image border and in more complex situations. The HD map created based on the semantic segmentation predicted by the neural network is only slightly worse, with a recall of 96\% and a precision of 96\%.
This indicates that the quality of semantic segmentation is high enough for the intended purpose.
In summary, the mapping results suggest that more than 95\% of the manual contour annotations can be done automatically. This drastically reduces the effort required to create maps from aerial imagery.

\begin{table}[htbp]
\caption{Map Creation Error}
\label{table:map_error}
\begin{center}
\begin{tabular}{ccc}
\hline
\textbf{Input} & \textbf{Precision}& \textbf{Recall} \\
\hline
reference (human annotations) & 0.97 & 0.97 \\
prediction (trained model outputs) & 0.96 & 0.96 \\ 
\hline
\end{tabular}
\label{tab3}
\end{center}
\end{table}

\section{CONCLUSIONS}

This paper is motivated by the increasing automation of road traffic, which has created a high demand for HD maps for the research, development, and operation of automated vehicles. Aerial imagery provides a readily available source of data compared to expensive sensor-equipped measurement vehicles. However, manually creating HD maps from aerial photographs is labour-intensive.

To address this challenge, we have proposed a method that combines neural networks for semantic segmentation with classical image processing algorithms to automatically generate prototype HD maps in the Lanelet2 format from aerial photographs. We created a dataset of 63 aerial photographs of urban street segments in Germany and semantically annotated them into eight relevant classes to train the network.

To achieve a large receptive field and highly effective segmentation resolution simultaneously, we made modifications to the UPerNet architecture. This allowed us to achieve a mean Intersection over Union (mIoU) of 70\% for segmenting the aerial images. Furthermore, we achieved 95\% accuracy in classifying road arrows and a contour recall of 91\% compared to the human reference for creating the HD map.

By exporting the map to the Lanelet2 format, the automatically created prototype can be easily extended with application-specific information. This demonstrates the potential of semi-automated processing of aerial imagery for HD map creation, as a significant proportion of the labour-intensive steps can be automated.

\addtolength{\textheight}{-12cm}   









\bibliographystyle{ieeetr} 
\bibliography{bibliography}

\begin{thebibliography}{10}

\bibitem{sae}
{The Society of Automotive Engineers International}, ``Taxonomy and definitions
  for terms related to driving automation systems for on-road motor vehicles,''
  2016.

\bibitem{liu_wang_zhang_2020}
R.~Liu, J.~Wang, and B.~Zhang, ``High definition map for automated driving:
  Overview and analysis,'' {\em Journal of Navigation}, vol.~73, no.~2, 2020.

\bibitem{s18103270}
H.~Cai, Z.~Hu, G.~Huang, D.~Zhu, and X.~Su, ``Integration of gps, monocular
  vision, and high definition (hd) map for accurate vehicle localization,''
  {\em Sensors}, vol.~18, no.~10, 2018.

\bibitem{poggenhans2018precise}
F.~Poggenhans, N.~O. Salscheider, and C.~Stiller, ``Precise localization in
  high-definition road maps for urban regions,'' in {\em 2018 IEEE/RSJ
  international conference on intelligent robots and systems (IROS)},
  pp.~2167--2174, 2018.

\bibitem{yoon2021high}
Y.~Yoon, H.~Chae, and K.~Yi, ``High-definition map based motion planning, and
  control for urban autonomous driving,'' tech. rep., Society of Automotive
  Engineers, 2021.

\bibitem{yoneda2020robust}
Y.Keisuke, K.~Akisuke, S.~Naoki, A.~Toru, A.~Mohammad, and Y.~Ryo, ``Robust
  traffic light and arrow detection using digital map with spatial prior
  information for automated driving,'' {\em Sensors}, vol.~20, no.~4, p.~1181,
  2020.

\bibitem{crowd}
L.Martin, J.~Dominik, S.~Julian, P.~David, and H.~Andreas, ``Crowdsourced hd
  map patches based on road model inference and graph-based slam,'' in {\em
  IEEE Intell. Veh. Symp. (IV)}, 2019.

\bibitem{xu2021topo}
Z.~Xu, Y.~Sun, and M.~Liu, ``Topo-boundary: A benchmark dataset on topological
  road-boundary detection using aerial images for autonomous driving,'' {\em
  IEEE Robotics and Automation Letters}, vol.~6, no.~4, 2021.

\bibitem{josm}
M.~Haklay and P.~Weber, ``Openstreetmap: User-generated street maps,'' {\em
  IEEE Pervasive computing}, vol.~7, no.~4, 2008.

\bibitem{vargas2020openstreetmap}
J.~Vargas-Munoz, S.~Srivastava, D.~Tuia, and A.~Falcao, ``Openstreetmap:
  Challenges and opportunities in machine learning and remote sensing,'' {\em
  IEEE Geoscience and Remote Sensing Magazine}, vol.~9, no.~1, 2020.

\bibitem{lanelet2}
P.~Fabian, P.~Jan-Hendrik, J.~Johannes, O.~Stefan, N.~Maximilian, K.~Florian,
  and M.~Matthias, ``Lanelet2: A high-definition map framework for the future
  of automated driving,'' in {\em IEEE Intell. Trans. Sys. Conf. (ITSC)}, 2018.

\bibitem{racq2020}
S.~Hausler and M.~Milford, ``Map creation, monitoring and maintenance for
  automated driving – literature review,'' tech. rep., 2020.

\bibitem{liu2020creating}
S.~Liu, L.~Li, J.~Tang, S.~Wu, and J.-L. Gaudiot, ``Creating autonomous vehicle
  systems,'' {\em Synthesis Lectures on Computer Science}, vol.~8, no.~2, 2020.

\bibitem{crowds}
M.~Liebner, D.~Jain, J.~Schauseil, D.~Pannen, and A.~Hackelöer, ``Crowdsourced
  hd map patches based on road model inference and graph-based slam,'' in {\em
  IEEE Intell. Veh. Symp. (IV)}, 2019.

\bibitem{8768011}
M.~Metwally, T.~M. Bazan, and F.~Eltohamy, ``Design of very high-resolution
  satellite telescopes part i: Optical system design,'' {\em IEEE Transactions
  on Aerospace and Electronic Systems}, vol.~56, no.~2, 2020.

\bibitem{c3}
H.~Jiuxiang, R.~Anshuman, F.~J. C., C.~Ming, and W.~Peter, ``Road network
  extraction and intersection detection from aerial images by tracking road
  footprints,'' {\em IEEE Transactions on Geoscience and Remote Sensing},
  vol.~45, no.~12, 2007.

\bibitem{deeproadmapper}
G.~Máttyus, W.~Luo, and R.~Urtasun, ``Deeproadmapper: Extracting road topology
  from aerial images,'' in {\em 2017 IEEE International Conference on Computer
  Vision (ICCV)}.

\bibitem{kim2006efficient}
J.~Kim, D.~Han, K.~Yu, Y.~Kim, and S.~Rhee, ``Efficient extraction of road
  information for car navigation applications using road pavement markings
  obtained from aerial images,'' {\em Canadian Journal of Civil Engineering},
  vol.~33, no.~10, 2006.

\bibitem{azimi2018aerial}
S.~Azimi, P.~Fischer, M.~Korner, and P.~Reinartz, ``Aerial lanenet:
  Lane-marking semantic segmentation in aerial imagery using wavelet-enhanced
  cost-sensitive symmetric fully convolutional neural networks,'' {\em IEEE
  Transactions on Geoscience and Remote Sensing}, vol.~57, no.~5, 2019.

\bibitem{markscaps}
Y.~Yu, Y.~Li, C.~Liu, J.~Wang, C.~Yu, X.~Jiang, L.~Wang, Z.~Liu, and Y.~Zhang,
  ``Markcapsnet: Road marking extraction from aerial images using
  self-attention-guided capsule network,'' {\em IEEE Geoscience and Remote
  Sensing Letters}, vol.~19, 2022.

\bibitem{6308139}
L.~Zhang, F.~Xiang, J.~Pu, and Z.~Zhang, ``Application of improved hu moments
  in object recognition,'' in {\em 2012 IEEE International Conference on
  Automation and Logistics}, 2012.

\bibitem{c10}
O.~Ronneberger, P.~Fischer, and T.~Brox, ``U-net: Convolutional networks for
  biomedical image segmentation,'' in {\em International Conference on Medical
  image computing and computer-assisted intervention}, 2015.

\bibitem{lin2017feature}
T.~Xiao, Y.~Liu, B.~Zhou, Y.~Jiang, and J.~Sun, ``Unified perceptual parsing
  for scene understanding,'' in {\em European Conf. on Comput. Vision (ECCV)},
  2018.

\bibitem{c11}
T.~Lin, P.~Doll, R.~Girshick, K.~He, B.~Hariharan, and S.~Belongie, ``Feature
  pyramid networks for object detection,'' in {\em IEEE Conf. on Comput. vision
  and pattern recognition}, 2017.

\bibitem{c13}
H.~Zhao, J.~Shi, X.~Qi, X.~Wang, and J.~Jia, ``Pyramid scene parsing network,''
  in {\em IEEE Conf. on Comput. vision and pattern recognition}, 2017.

\bibitem{c14}
K.~He, X.~Zhang, S.~Ren, and J.~Sun, ``Deep residual learning for image
  recognition,'' in {\em IEEE Conf. on Comput. vision and pattern recognition},
  2016.

\bibitem{devries2017improved}
T.~DeVries and G.~W. Taylor, ``Improved regularization of convolutional neural
  networks with cutout,'' {\em arXiv:1708.04552}, 2017.

\end{thebibliography}

\end{document}